\documentclass[lettersize,journal,twoside]{IEEEtran}
\usepackage{amsmath,amsfonts}
\usepackage{algorithm}
\usepackage{array}
\usepackage{textcomp}
\usepackage{stfloats}
\usepackage{url}
\usepackage{verbatim}
\usepackage{graphicx}
\usepackage{cite}
\usepackage{graphics}           
\usepackage{times}              
\usepackage{amsmath}            
\usepackage{amssymb}            
\usepackage{graphicx}
\usepackage{algorithm}
\usepackage[noend]{algpseudocode}
\usepackage{booktabs}
\usepackage{color}
\usepackage{multirow}
\usepackage{subcaption}
\usepackage{rotating}
\usepackage{cite}
\usepackage{makecell}
\usepackage[table,xcdraw]{xcolor}
\usepackage{colortbl}
\usepackage{hyperref}
\usepackage{soul}

\def\thline{\Xhline{2.3\arrayrulewidth}}
\definecolor{Gray}{gray}{0.85} 
\definecolor{Tgray}{gray}{0.85} 
\sethlcolor{Tgray}
\def\secref#1{Section~\ref{#1}}
\def\figref#1{Fig.~\ref{#1}}
\def\subfigref#1#2{Fig.~\ref{#1}(#2)}
\def\tabref#1{Table~\ref{#1}}
\def\eqref#1{(\ref{#1})}

\def\vstabcap{\vspace{-0.16cm}}
\def\vstabfoot{\vspace{0.07cm}}
\def\vstabstart{\vspace{-0.35cm}}

\def\vsfigcap{\vspace{-0.12cm}}

\newcommand{\rom}[1]{\uppercase\expandafter{\romannumeral #1\relax}}
\newcommand{\romsmall}[1]{\lowercase\expandafter{\romannumeral #1\relax}}
\makeatletter
\usepackage{xspace}
\DeclareRobustCommand\onedot{\futurelet\@let@token\@onedot}
\def\@onedot{\ifx\@let@token.\else.\null\fi\xspace}
\def\etal{{\textit{et al}}\onedot}

\def\ie{i.e\onedot}

\makeatother
\usepackage{array}
\newcolumntype{L}[1]{>{\raggedright\let\newline\\\arraybackslash\hspace{0pt}}m{#1}}
\newcolumntype{C}[1]{>{\centering\let\newline\\\arraybackslash\hspace{0pt}}m{#1}}
\newcolumntype{R}[1]{>{\raggedleft\let\newline\\\arraybackslash\hspace{0pt}}m{#1}}

\def\transpose{\mathsf{T}}
\def\gyro{\boldsymbol{\omega}}
\def\accel{\mathbf{a}}
\def\biasgyro{\mathbf{b}\gyro}
\def\biasaccel{\mathbf{b}\accel}
\def\normalvector{{ }^G \mathbf{e}}

\newcommand{\dataset}[1]{\scalebox{0.8}[1]{\texttt{#1}}}
\newcommand{\datasettable}[1]{\scalebox{0.85}[1]{\texttt{#1}}}

\begin{document}

\title{LODESTAR: Degeneracy-Aware LiDAR-Inertial Odometry with Adaptive Schmidt-Kalman Filter and Data Exploitation}

\author{Eungchang Mason Lee, \textit{Member, IEEE}, Kevin Christiansen Marsim, \textit{Graduate Student Member, IEEE},\\and Hyun Myung, \textit{Senior Member, IEEE}%
    \thanks{Manuscript received: July, 14, 2025; Revised October, 11, 2025; Accepted November, 1, 2025.
    This paper was recommended for publication by Editor Sven Behnke upon evaluation of the Associate Editor and Reviewers' comments.
    (\textit{Corresponding author: Hyun Myung.})}
    \thanks{Eungchang Mason Lee is with KAIST InnoCORE LLM, KAIST, Daejeon, 34141, Republic of Korea (e-mail: eungchang\_mason@kaist.ac.kr).}
    \thanks{Kevin Christiansen Marsim and Hyun Myung are with the School of Electrical Engineering, KAIST, Daejeon, 34141, Republic of Korea (e-mail: kevinmarsim@kaist.ac.kr; hmyung@kaist.ac.kr).}
    \thanks{Digital Object Identifier (DOI): see top of this page.}
}

\markboth{IEEE Robotics and Automation Letters. Preprint Version. Accepted November, 2025}%
{Lee \MakeLowercase{\textit{et al.}}: LODESTAR: Degeneracy-Aware LiDAR-Inertial Odometry with Adaptive Schmidt-Kalman Filter and Data Exploitation}


\maketitle

\begin{abstract}
    LiDAR-inertial odometry (LIO) has been widely used in robotics due to its high accuracy.
    However, its performance degrades in degenerate environments, such as long corridors and high-altitude flights, where LiDAR measurements are imbalanced or sparse, leading to ill-posed state estimation.
    In this letter, we present LODESTAR, a novel LIO method that addresses these degeneracies through two key modules: degeneracy-aware adaptive Schmidt-Kalman filter (DA-ASKF) and degeneracy-aware data exploitation (DA-DE).
    DA-ASKF employs a sliding window to utilize past states and measurements as additional constraints.
    Specifically, it introduces degeneracy-aware sliding modes that adaptively classify states
    as active or fixed based on their degeneracy level.
    Using Schmidt-Kalman update, it partially optimizes active states while preserving fixed states.
    These fixed states influence the update of active states via their covariances,
    serving as reference anchors--akin to a lodestar.
    Additionally, DA-DE prunes less-informative measurements from active states
    and selectively exploits measurements from fixed states, based on their localizability contribution and the condition number of the Jacobian matrix.
    Consequently, DA-ASKF enables degeneracy-aware constrained optimization and
    mitigates measurement sparsity, while DA-DE addresses measurement imbalance.
    Experimental results show that LODESTAR outperforms existing LiDAR-based odometry
    methods and degeneracy-aware modules in terms of accuracy and robustness
    under various degenerate conditions.
\end{abstract}

\begin{IEEEkeywords}
Localization, Mapping, SLAM, Degeneracy.
\end{IEEEkeywords}

\section{Introduction}\label{sec:introduction}
\IEEEPARstart{M}{obile} robots have been increasingly deployed to autonomously perform tasks in place of humans, including search and rescue, exploration, mapping, and structural inspection.
These tasks often require mobile robots to operate in indoor spaces, high-altitude environments, or unstructured terrains,
where accurate and robust self-localization becomes essential for reliable operation and navigation~\cite{tuna2025tfr,ebadi2023tro-darpa-slam,lee2021iros-real,shan2020iros-liosam,vizzo2023ral-kissicp,tuna2023tro-xicp,chen2022ral-dlo,jeon2021ral-run,kim2022ral-step,lim2024ijrr-quatro++}.

\begin{figure}[t]
    \centering
    \includegraphics[width=0.46\textwidth]{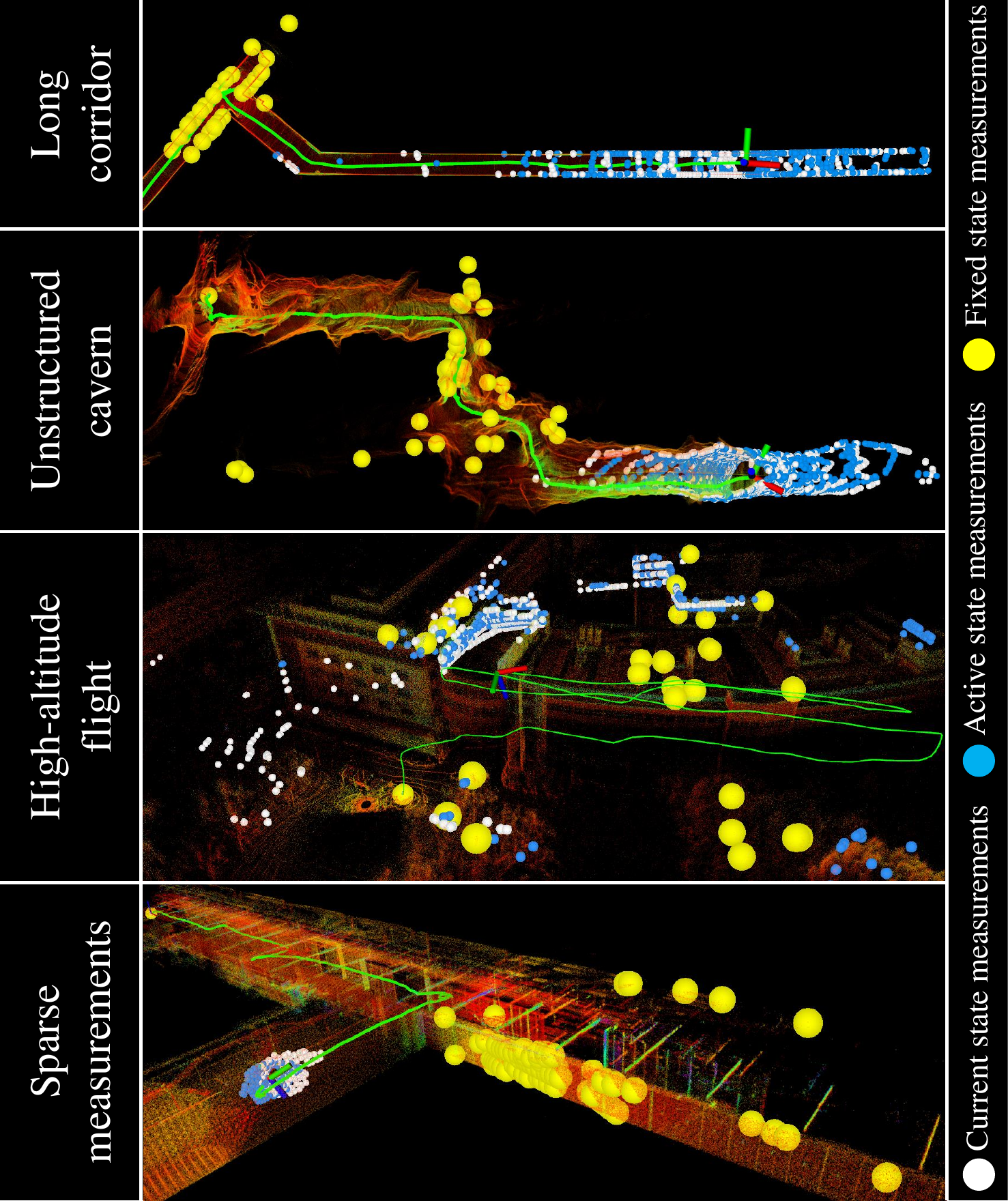}
    \\ \vspace{-0.07cm}
    \caption{
    Instances of LODESTAR under various degenerate conditions.
    In degenerate environments, LiDAR measurements are often sparse or imbalanced.
    LODESTAR mitigates these problems using DA-ASKF and DA-DE, which selectively utilize past states and measurements.
    States are classified into active and fixed based on their degeneracy level.
    Measurements from current and active states are pruned based on their localizability contribution.
    Then, measurements from fixed states in degenerate directions are exploited to resolve measurement imbalance.
    }
    \label{fig:figure1}
    \vspace{-0.2cm}
\end{figure}

In particular, among various sensors, LiDAR has been steadily adopted due to its precise distance measurements,
as a core sensor for odometry estimation such as LiDAR-only odometry (LO) and LiDAR-inertial odometry (LIO),
the latter of which additionally utilizes IMUs~\cite{shan2020iros-liosam, xu2022tro-fastlio2, jiang2022iros-pumalio, vizzo2023ral-kissicp, wang2023ral-swlio, chen2024ral-iglio, chen2022ral-dlo, lim2023ur-adalio, chen2023icra-dlio, huang2024iros-lalio, yang2024sensors-naloam, chung2024ral-nvliom, lee2024ral-genzicp, tuna2023tro-xicp}.

Despite the highly accurate distance measurements, LiDAR-based odometry can easily fail under degenerate conditions,
such as long corridors, unstructured caverns, and high-altitude flights, where LiDAR measurements are imbalanced or sparse--meaning that scanned points are unevenly distributed across directions (imbalance) or insufficient in quantity (sparsity).
These conditions result in ill-posed estimation, causing odometry drift or divergence, which can lead to mission failure~\cite{tuna2025tfr, lim2023ur-adalio, tuna2023tro-xicp, ebadi2023tro-darpa-slam, chen2023icra-dlio, huang2024iros-lalio, lee2024ral-genzicp, chung2024ral-nvliom, yang2024sensors-naloam, zhang2016icra-zhang, hinduja2019iros-hinduja, gelfand2003threedim-degeneracy}.
To overcome these degeneracies, various methods have been proposed, including degeneracy-aware constrained optimization to prevent the divergence~\cite{tuna2023tro-xicp,zhang2016icra-zhang,hinduja2019iros-hinduja},
LiDAR data weight adjustment to handle measurement imbalance~\cite{huang2024iros-lalio,chung2024ral-nvliom,yang2024sensors-naloam},
and data augmentation to alleviate measurement sparsity~\cite{chen2022ral-dlo,lim2023ur-adalio,chen2023icra-dlio,lee2024ral-genzicp}.
However, these methods often address only one aspect of the degeneracy and thus may not be sufficient to handle various degenerate conditions.

In this paper, we propose \textbf{LODESTAR}, \textbf{L}iDAR-inertial \textbf{O}dometry with \textbf{D}ata \textbf{E}xploitation and \textbf{S}chmid\textbf{T}-Kalman \textbf{A}daptive filte\textbf{R},
a novel LIO method that simultaneously addresses measurement imbalance, measurement sparsity, and degeneracy-aware constrained optimization,
resulting in enhanced robustness and accuracy.
LODESTAR builds upon the error-state iterative Kalman filter (ESIKF) framework~\cite{xu2022tro-fastlio2} and proposes two key modules:
(\romsmall{1}) degeneracy-aware adaptive Schmidt-Kalman filter (DA-ASKF) that employs Schmidt-Kalman update~\cite{schmidt1966acs-schmidt} within a sliding window to adaptively utilize past states and measurements;
(\romsmall{2}) degeneracy-aware data exploitation (DA-DE) that selectively exploits measurements based on their localizability contribution.
Instances of LODESTAR in various degenerate scenarios are shown in \figref{fig:figure1}.
The main contributions of this paper are as follows:
\begin{itemize}
    \item \textbf{DA-ASKF} that introduces adaptive degeneracy-aware sliding modes within a sliding window,
    classifying past states into active and fixed based on their degeneracy level,
    and leveraging past states and measurements as additional constraints to enhance robustness and accuracy.
    Through Schmidt-Kalman update, these fixed states guide updates of current and active states via their covariances,
    serving as reference anchors--akin to a lodestar.
    \item \textbf{DA-DE} that selectively exploits LiDAR measurements based on their localizability contribution.
    Guided by the condition number of the Jacobian matrix, it incrementally exploits data from current, active, and fixed states
    until the optimization becomes numerically stable.
    \item \textbf{Extensive evaluation} on multiple degenerate datasets, including imbalanced and sparse measurements from long corridors, unstructured caverns, and high-altitude flights.
    LODESTAR demonstrates improved or comparable accuracy and robustness relative to state-of-the-art LiDAR-based odometry methods and degeneracy-aware modules.
    \item \textbf{Open-source release and reproducibility}
    including the full source code, experimental configurations, and our self-implemented baselines to facilitate reproducibility.
\end{itemize}

\section{Related Works}\label{sec:related-works}
LiDAR-based odometry estimates the relative transformation between consecutive scans mainly using the point cloud registration methods, such as point-to-point~\cite{besl1992point-to-point} and point-to-plane~\cite{rusinkiewicz2001point-to-plane}.
These methods are typically formulated as nonlinear optimizations, where the objective is to minimize distances between the measured points and the correspondences from prior scans or the global map.
The nonlinear optimization is typically solved using iterative methods, such as Gauss-Newton or Levenberg-Marquardt methods,
where an inverse of Jacobian must be computed to update the state~\cite{nocedal1999gn-lm-gradient}.
When measurements are sparse or imbalanced, Jacobian becomes (nearly) singular,
leading to numerical instability or divergence, \ie,~degeneracy~\cite{tuna2023tro-xicp, zhang2016icra-zhang, hinduja2019iros-hinduja, gelfand2003threedim-degeneracy}.
Under degeneracy, Hessian spectrum becomes highly anisotropic, flattening certain directions;
state updates become slower and more sensitive~\cite{tuna2025tfr}.

Therefore, degeneracy is commonly quantified by either the minimum singular value~\cite{chung2024ral-nvliom,zhang2016icra-zhang,tuna2023tro-xicp}, 
or the condition number of the Jacobian~\cite{yang2024sensors-naloam,lee2024ral-genzicp,hinduja2019iros-hinduja}, which is defined as the ratio of the largest to the smallest singular value~\cite{cheney1998condition-number}.

In this work, we adopt the condition number as a practical indicator rather than an absolute measure of solvability;
a high condition number does not always imply failure,
yet in real-time LIO, it reliably captures both sparsity and imbalance of measurements,
where additional constraints can be beneficial.
We are not focusing on an optimal solution, but try to provide a practical solution for real-time LIO.
We detect degeneracy early and adapt updates to keep the filter stable during short-term low-information intervals with non-zero returns,
acknowledging that prolonged zero-return intervals still cause IMU dead-reckoning drift as in most LIO systems.

Consistent with this practical stance, existing degeneracy-aware odometry methods can be classified into:
(\romsmall{1}) degeneracy-aware constrained optimization; (\romsmall{2}) imbalance-aware weight adjustment;
and (\romsmall{3}) sparsity-aware data augmentation. 

\textbf{Degeneracy-aware constrained optimization:}
These methods add constraints to improve robustness to degeneracy.  
Zhang~\etal~\cite{zhang2016icra-zhang} and Hinduja~\etal~\cite{hinduja2019iros-hinduja} constrain state updates along degenerate directions to retain their previous values.
X-ICP~\cite{tuna2023tro-xicp} classifies directions by localizability, updating states in localizable directions only, and holding states in non-localizable directions fixed;
states in partially localizable directions are updated using additional constraints from resampled measurements.
However, these methods forgo updates in degenerate directions,
causing estimation errors to accumulate over time, potentially resulting in drift or divergence.
 
\textbf{Imbalance-aware weight adjustment:}
These methods adjust weights or covariances to compensate for measurements imbalance.  
LA-LIO~\cite{huang2024iros-lalio} scales measurement covariance based on the number of points within local patches.  
NA-LOAM~\cite{yang2024sensors-naloam} performs singular value decomposition on the measurement matrix, scaling weights by the inverse of the maximum singular value.
NV-LIOM~\cite{chung2024ral-nvliom} modifies covariances in pose graph optimization according to the singular values of Jacobian.

\textbf{Sparsity-aware data augmentation:}
These methods mitigate measurement sparsity by augmenting the scan or reference map for registration.
DLO~\cite{chen2022ral-dlo} and DLIO~\cite{chen2023icra-dlio} adaptively spawn reference frames based on environment spaciousness.
AdaLIO~\cite{lim2023ur-adalio} adjusts the downsampling resolution of scans according to the number of points scanned at each frame.

However, both categories above focus on either imbalance or sparsity, respectively, and do not guarantee robustness in degenerate environments.
GenZ-ICP~\cite{lee2024ral-genzicp} addresses both imbalance and sparsity by combining point-to-point and point-to-plane registration with weight adjustment.  
However, it does not utilize measurements from past states at all.

\textbf{Our distinction:}
To our knowledge, LODESTAR is the first framework that simultaneously addresses degeneracy-aware constrained optimization, measurement sparsity, and imbalance.
Unlike prior works, LODESTAR updates current and active states in both degenerate and non-degenerate directions, using fixed states as reference anchors.
Its sliding window-based approach utilizes past measurements to mitigate sparsity, while the selective data exploitation resolves imbalance.
Note that a few methods adopt a sliding window on ESIKF.
SW-LIO~\cite{wang2023ral-swlio} updates all states using all available measurements. 
PUMA-LIO~\cite{jiang2022iros-pumalio} instead fixes all past states. 
However, both methods address neither degeneracy nor data exploitation.

\section{LODESTAR: Degeneracy-Aware LiDAR-Inertial Odometry}\label{sec:methodology}

\begin{figure*}[t]
    \centering
    \begin{subfigure}[c]{0.218\textwidth}
        \centering
        \includegraphics[width=\textwidth]{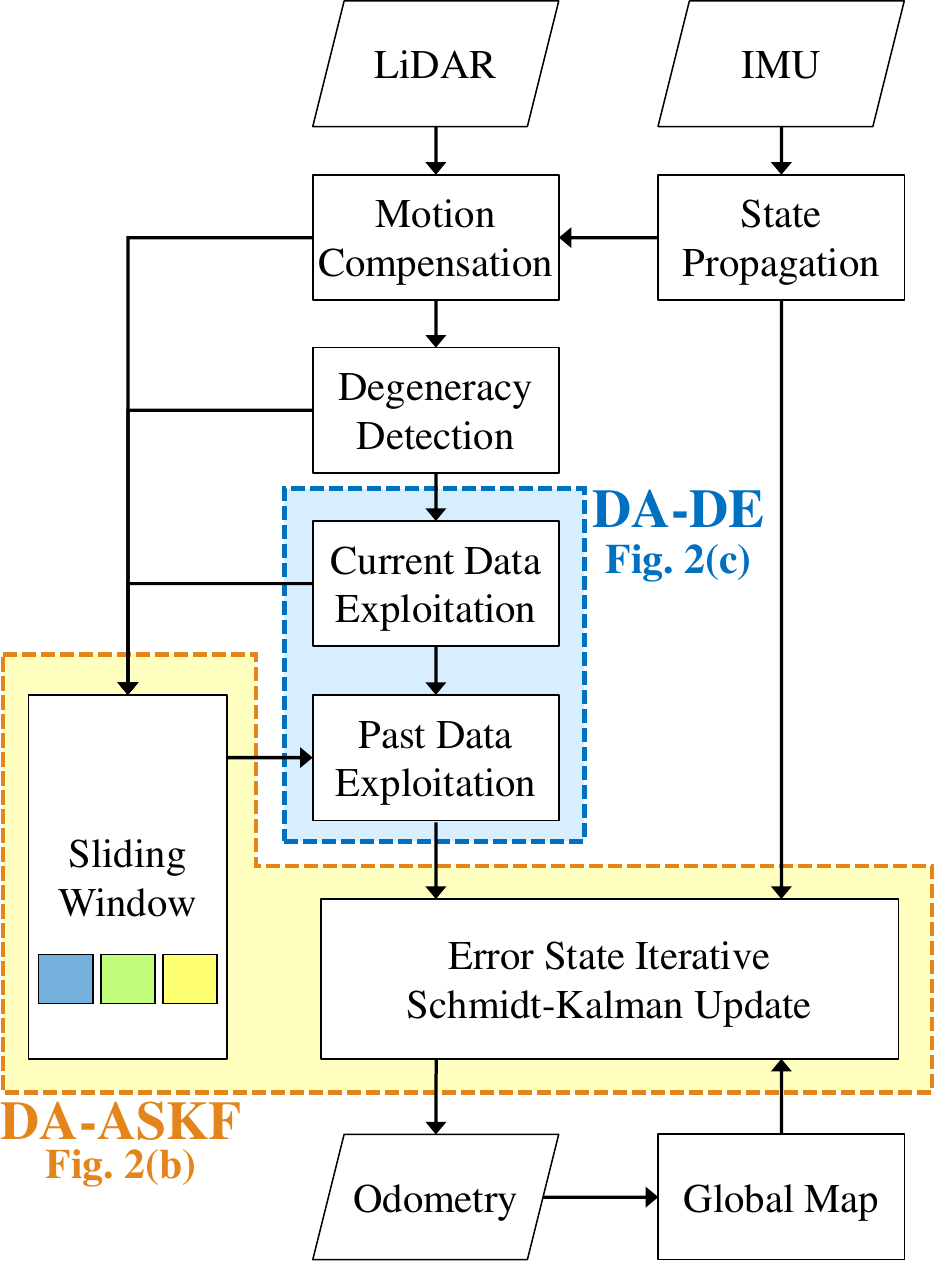}
        \captionsetup{skip=2pt}
        \caption{}
    \end{subfigure}
    \hspace{0.5cm}
    \begin{subfigure}[c]{0.528\textwidth}
        \centering
            \includegraphics[width=\textwidth]{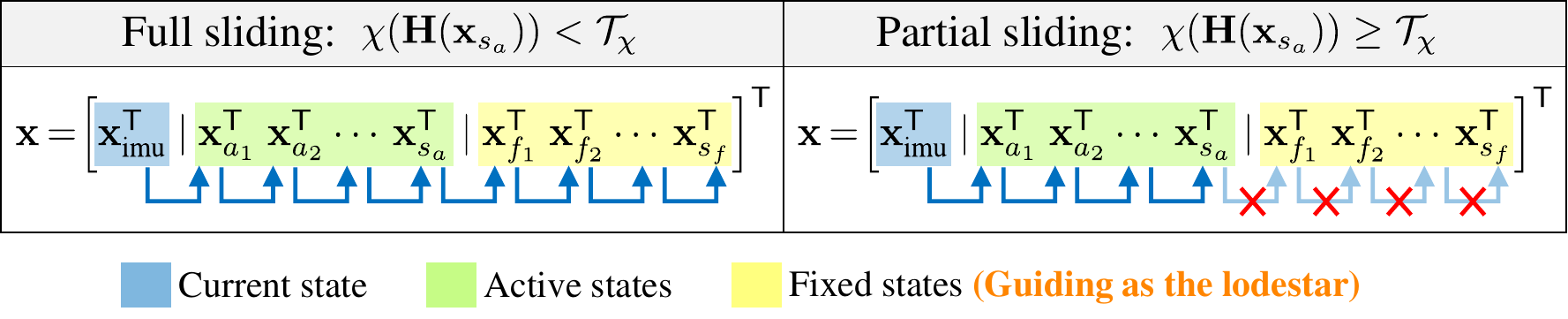}
            \captionsetup{skip=1pt}
            \caption{}
        
            \vspace{0.3em}
            
            \includegraphics[width=\textwidth]{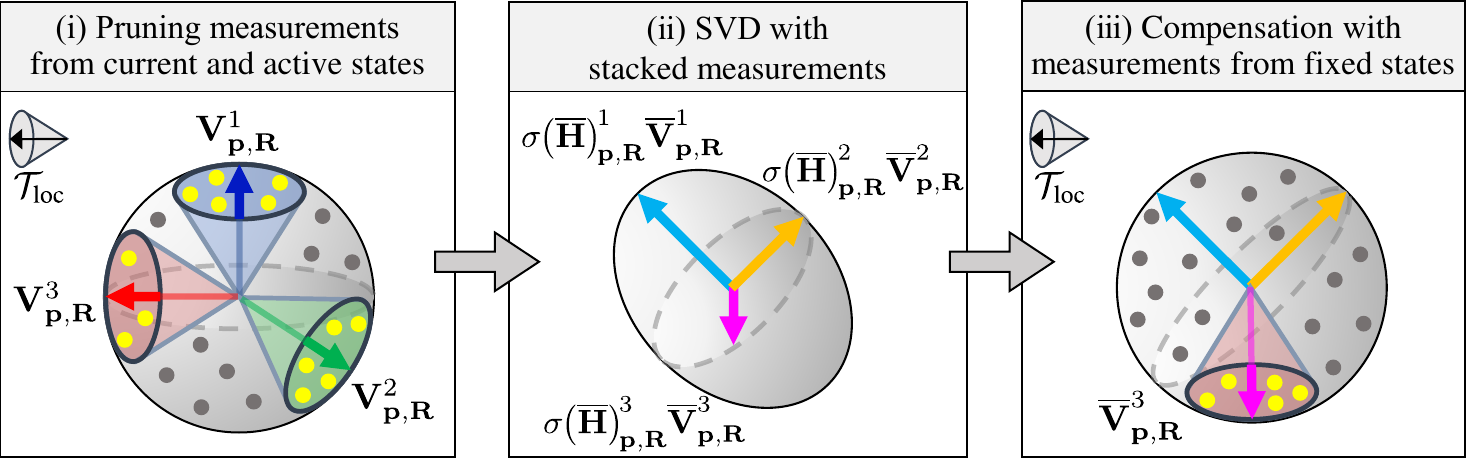}
            \captionsetup{skip=2pt}
            \caption{}
    \end{subfigure}
    \vspace{-0.15cm}
    \caption{Overview and core components of the proposed LODESTAR.
    \textbf{(a)} Flowchart of LODESTAR.
    \textbf{(b)} Degeneracy-aware sliding modes in DA-ASKF (degeneracy-aware adaptive Schmidt-Kalman filter).
    \textbf{(c)} Selective data pruning and compensation in DA-DE (degeneracy-aware data exploitation).}
    \label{fig:lodestar}
    \vspace{-0.3cm}
\end{figure*}

\subsection{Overview}\label{sec:overview}
The overview of the proposed \textbf{LODESTAR} is illustrated in~\subfigref{fig:lodestar}{a}.
Following the ESIKF framework~\cite{xu2022tro-fastlio2},
IMU measurements are first used to propagate the state and compensate for motion distortion in LiDAR scans.
The degeneracy level of the current state is evaluated from the LiDAR scans.
DA-DE selectively exploits measurements associated with the current and past states within the sliding window.
The selected data is then used to iteratively update current and active error states through Schmidt-Kalman update in DA-ASKF,
while fixed states contribute to the estimation as reference anchors.
After the update, the current and active states are shifted by one frame,
and the last active state is removed or transferred to the fixed state set, depending on its degeneracy level.


\subsection{Filter Design}\label{sec:filter-design}
\begin{table}[b]
    \vstabstart
    \centering
    \caption{Notation of States and Variables}\label{tab:notation}
    \vstabcap
    \scriptsize
    \renewcommand{\arraystretch}{1.07}
    \setlength\tabcolsep{4.5pt}
    \begin{tabular}{ll}
        \thline
        Symbol & Description \\
        \hline        
        $\mathbf{x}, \widehat{\mathbf{x}}, \overline{\mathbf{x}}$ & True, propagated, and updated states, respectively. \rule{0pt}{1.0em}\\
        $\widetilde{\mathbf{x}}$ & Error state defined as $\widetilde{\mathbf{x}}=\mathbf{x}\boxminus\widehat{\mathbf{x}}$. \\
        ${}^{I}\mathbf{x}_i$, ${}^{L}\mathbf{x}_k$ & States at the $i$-th IMU input and $k$-th LiDAR scan, respectively. \\
        ${}^{L}\mathbf{x}_k^\lambda$ & State at the $\lambda$-th iteration of the $k$-th LiDAR scan. \\
        ${}^G(\cdot), {}^L(\cdot), {}^I(\cdot)$ & Variables in global, LiDAR, and IMU frames, respectively. \\
        \thline
    \end{tabular}
\end{table}
    
In this subsection, we briefly revisit the ESIKF~\cite{xu2022tro-fastlio2} and highlight our degeneracy-aware extensions.
Notations are summarized in \tabref{tab:notation}.
Assuming the global frame is aligned with the first IMU frame, and the extrinsic transformation between the LiDAR and IMU frames is known,
the full state $\mathbf{x}$ on the compound manifold $\boldsymbol{\mathcal{M}}$ consists of the current IMU state, active past states, and fixed past states as follows:
\begin{equation}\label{eq:state}
	\begin{aligned}
        \mathbf{x} & \triangleq 
            \left[\mathbf{x}_{\text{imu}}^{\transpose} \
            \mathbf{x}_{a_1}^{\transpose} \ \mathbf{x}_{a_2}^{\transpose} \cdots \, \mathbf{x}_{s_a}^{\transpose} \
            \mathbf{x}_{f_1}^{\transpose} \ \mathbf{x}_{f_2}^{\transpose} \cdots \, \mathbf{x}_{s_f}^{\transpose} \right]^{\transpose} \in \boldsymbol{\mathcal{M}},\\
        \mathbf{x}_{\text{imu}} & \triangleq
            \left[ {}^G\mathbf{R}_I^{\transpose} \ \ {}^G\mathbf{p}_I^{\transpose} \ \ {}^G\mathbf{v}_I^{\transpose} \ \
            \mathbf{b}_{\gyro}^{\transpose} \ \ \mathbf{b}_{\accel}^{\transpose} \ \ {}^G\mathbf{g}^{\transpose}
            \right]^{\transpose}, \\
        \boldsymbol{\mathcal{M}} & \triangleq SO(3) \times \mathbb{R}^{15} \times\left[SO(3) \times \mathbb{R}^3\right]^{s_w}, \\
        \mathbf{u} & \triangleq \left[\gyro^{\transpose} \ \ \accel^{\transpose} \right]^{\transpose}, \ \
        \mathbf{w} \triangleq \left[\mathbf{n}_{\gyro}^{\transpose} \ \ \mathbf{n}_{\accel}^{\transpose} \ \
        \mathbf{n}_{\biasgyro}^{\transpose} \ \ \mathbf{n}_{\biasaccel}^{\transpose} \right]^{\transpose},
    \end{aligned}
\end{equation}
\noindent where ${}^G\mathbf{R}_I$, ${}^G\mathbf{p}_I$, ${}^G\mathbf{v}_I$, $\mathbf{b}_{\gyro}$, $\mathbf{b}_{\accel}$, ${}^G\mathbf{g}$
denote the IMU rotation, position, velocity in the global frame, gyroscope bias, accelerometer bias, and gravity vector, respectively.
The IMU input $\mathbf{u}$ consists of the angular rate $\gyro$ and acceleration $\accel$, while the noise $\mathbf{w}$ includes
their corresponding noises $\mathbf{n}_{\gyro}$, $\mathbf{n}_{\accel}$ and bias noises $\mathbf{n}_{\biasgyro}$, $\mathbf{n}_{\biasaccel}$.
$\mathbf{x}_{a_i}, \mathbf{x}_{f_i}\!\triangleq\! \left[ {}^G\mathbf{R}_{I_i}^{\transpose} \ {}^G\mathbf{p}_{I_i}^{\transpose} \right]^{\transpose}$ are the active and fixed states, respectively.
$s_a$ and $s_f$ are the numbers of active and fixed states; $s_w\!=\!s_a+s_f$ is the window size.

Between consecutive LiDAR scans, the state is propagated with the IMU input $\mathbf{u}_i$ and the noise $\mathbf{w}_i$
based on the IMU's kinematic model as follows:
\begin{align}
    {}^{I}\widehat{\mathbf{x}}_{i+1}&={}^{I}\widehat{\mathbf{x}}_i \boxplus\left(\mathbf{f}\left({}^{I}\widehat{\mathbf{x}}_i, \mathbf{u}_i, \mathbf{0}\right) \Delta t_i\right); \ \ {}^{I}\widehat{\mathbf{x}}_0={}^{L}\overline{\mathbf{x}}_{k\!-\!1}, \nonumber \\
    \widehat{\mathbf{P}}_{i+1}&=\mathbf{F}_{\widetilde{\mathbf{x}}_i} \widehat{\mathbf{P}}_i \mathbf{F}_{\widetilde{\mathbf{x}}_i}^{\transpose}+\mathbf{F}_{\mathbf{w}_i} \mathbf{Q} \mathbf{F}_{\mathbf{w}_i}^{\transpose}; \ \ \widehat{\mathbf{P}}_0=\overline{\mathbf{P}}_{k\!-\!1}, \nonumber \\
    \mathbf{f}({}^{I}\mathbf{x}_i, \mathbf{u}_i, \mathbf{w}_i) &= 
    \left[\begin{array}{c}
    \gyro_{i} - \mathbf{b}_{\gyro_i} - \mathbf{n}_{\gyro_i} \\
    {}^{G}\mathbf{v}_{I_i} \\
    {}^{G}\mathbf{R}_{I_i} \left( \accel_{i} - \mathbf{b}_{\accel_i} - \mathbf{n}_{\accel_i} \right) + {}^{G}\mathbf{g}_i \\
    \mathbf{n}_{\biasgyro_i} \\
    \mathbf{n}_{\biasaccel_i} \\
    \mathbf{0}_{\left(6s_w+3\right) \times 1}
    \end{array}\right],
    \label{eq:propagation}
\end{align}

\noindent where ${}^{I}\widehat{\mathbf{x}}_{i+1}$ and $\widehat{\mathbf{P}}_{i+1}$ are the propagated state and covariance,
$\mathbf{f}$ is the state transition function, $\Delta t_i$ is the IMU sample interval,
$\mathbf{F}_{\widetilde{\mathbf{x}}_i}$ and $\mathbf{F}_{\mathbf{w}_i}$ are Jacobians of $\mathbf{f}$ with respect to the error state ${}^{I}\widetilde{\mathbf{x}}_i$ and noise $\mathbf{w}_i$, respectively, as detailed in~\cite{xu2022tro-fastlio2}.
The operator $\boxplus$ denotes the addition on manifold~\cite{hertzberg2013infofusion-manifold}.

The error state ${}^{L}\widetilde{\mathbf{x}}_k^\lambda$ is iteratively estimated by solving the following maximum-a-posteriori (MAP) problem:
\begin{equation}\label{eq:maximum-a-posteriori}
	\min _{{}^{L}\widetilde{\mathbf{x}}_k^\lambda} \left( \lVert {}^{L}\mathbf{x}_k \boxminus {}^{L}\widehat{\mathbf{x}}_k\rVert_{\widehat{\mathbf{P}}_k}^2 + \sum_{j=1}^m \lVert \mathbf{z}_{k,j}^\lambda+\mathbf{H}_{k,j}^\lambda {}^{L}\widetilde{\mathbf{x}}_k^\lambda \rVert_{\mathbf{C}_j}^2 \right),
\end{equation}
\noindent where $\mathbf{z}^\lambda_{k,j}$ is the residual of the $j$-th LiDAR point at the $\lambda$-th iteration,
and $\mathbf{H}^\lambda_{k,j}$ is its Jacobian with respect to the state.
The operator $\boxminus$ denotes the subtraction on manifold~\cite{hertzberg2013infofusion-manifold}.
The $\lambda$-th iteration of the MAP problem is solved as follows:
\begin{align}
    \mathbf{P}&=\mathbf{J}^{-1} \widehat{\mathbf{P}}_{i+1}\mathbf{J}^{-\transpose}, \nonumber \\
    \mathbf{K}&=\left(\mathbf{H}^{\transpose} \mathbf{C}^{-1} \mathbf{H}+\mathbf{P}^{-1}\right)^{-1} \mathbf{H}^{\transpose} \mathbf{C}^{-1}, \label{eq:iteration} \\
    {}^{L}\widehat{\mathbf{x}}_k^{\lambda+1}&={}^{L}\widehat{\mathbf{x}}_k^\lambda \boxplus \left(-\mathbf{K} \mathbf{z}-(\mathbf{I}-\mathbf{K} \mathbf{H})\mathbf{J}^{-1}\left({}^{L}\widehat{\mathbf{x}}_k^\lambda \boxminus {}^{L}\widehat{\mathbf{x}}_k\right)\right), \nonumber
\end{align}
\noindent where $\mathbf{J}\!=\!\partial \left( \left({}^{L}\widehat{\mathbf{x}}_k^\lambda \boxplus {}^{L}\widetilde{\mathbf{x}}_k^\lambda\right) \boxminus {}^{L}\widehat{\mathbf{x}}_k \right) \big/ \partial {}^{L}\widetilde{\mathbf{x}}_k^\lambda$
is the corresponding Jacobian and $\mathbf{C}$ is the diagonal measurement noise covariance matrix.
$\mathbf{H}$ and $\mathbf{z}$ are stacked Jacobian and residual vectors, respectively.
Note that super- and subscripts $\left(\cdot\right)^\lambda_k$ of $\mathbf{P}$, $\mathbf{K}$, $\mathbf{H}$, $\mathbf{J}$ and $\mathbf{z}$
are omitted for simplicity.
After the maximum iteration or the convergence, the state is updated as follows:
\begin{equation}\label{eq:state-update}
    \begin{aligned}
        {}^{L}\overline{\mathbf{x}}_k&={}^{L}\widehat{\mathbf{x}}_k^{\lambda+1},\\
        \overline{\mathbf{P}}_k&=\left(\mathbf{I}-\mathbf{K} \mathbf{H}\right) \mathbf{P}\left(\mathbf{I}-\mathbf{H}^{\transpose} \mathbf{K}^{\transpose}\right)+\mathbf{K C K}^{\transpose}.
    \end{aligned}
\end{equation}

\noindent The details of Schmidt-Kalman update, measurement models with the past states, and sliding modes are explained in the following subsection.

\subsection{Degeneracy-Aware Adaptive Schmidt-Kalman Filter}\label{sec:da-askf}
\textbf{Measurement model with past states:}
Without additional coupling, the current state is independent from the previous measurements.
Hence, we adopt a change term $\Delta \mathbf{x}_k$ between adjacent states, inspired by SW-LIO~\cite{wang2023ral-swlio} as follows:
\begin{equation}\label{eq:change-term}
    \Delta \mathbf{x}_k \triangleq \left[\begin{array}{c}
    \widehat{\mathbf{R}}_{k} \\
    \widehat{\mathbf{p}}_{k}
    \end{array}\right] \boxminus\left[\begin{array}{c}
    \widehat{\mathbf{R}}_{{k\!-\!1}} \\
    \widehat{\mathbf{p}}_{{k\!-\!1}}
    \end{array}\right]
    = \left[\begin{array}{c}
        \Delta \widehat{\boldsymbol{\theta}}_{k} \\
        \Delta \widehat{\mathbf{p}}_{k}\end{array}\right] \in \mathbb{R}^6.
\end{equation}

\noindent Thus, the following equations can be formulated to couple the current and previous states:
\begin{equation}\label{eq:coupling}
	\begin{aligned}
    \mathbf{R}_{{k\!-\!1}}&=\mathbf{R}_{k}\left(\operatorname{Exp}\left(\Delta \widehat{\boldsymbol{\theta}}_k\right)\right)^{\transpose} \operatorname{Exp}\left(\delta \boldsymbol{\theta}_{k\!-\!1}\right), \\
    \mathbf{p}_{k\!-\!1}&=\mathbf{p}_k-\Delta \widehat{\mathbf{p}}_k+\delta \mathbf{p}_{k\!-\!1},
    \end{aligned}
\end{equation}

\noindent where $\operatorname{Exp}(\cdot)$ maps a 3D rotation vector to a rotation matrix on $SO(3)$~\cite{hertzberg2013infofusion-manifold}.
$\delta \boldsymbol{\theta}$ and $\delta \mathbf{p}$ are the rotation and position error states expressed in the tangent space of manifold, respectively.
Meanwhile, the point-to-plane residual $\mathbf{z}_{k,j}$ and the measurement function $\mathbf{h}(\cdot)$ are defined as follows:
\begin{equation}\label{eq:residual}
    \begin{aligned}
        \mathbf{h}_j\left(\mathbf{x}_{k},{ }^L \mathbf{n}_j\right) & \triangleq \normalvector_j^{{\transpose}} \left( \mathbf{R}_{k} \mathbf{D}_j + \mathbf{p}_{k} - { }^G \mathbf{q}_j \right), \\
        \mathbf{z}_{k,j} &= \mathbf{h}_j\left(\widehat{\mathbf{x}}_k, \mathbf{0}\right),\\
        \mathbf{D}_j &= { }^I \mathbf{R}_{L} ({ }^L\mathbf{d}_j + {}^L\mathbf{n}_j) + { }^I \mathbf{p}_{L},
    \end{aligned}
\end{equation}

\noindent where $\normalvector_j$ and ${}^L \mathbf{n}_j$ are the unit normal vector and noise of the $j$-th measurement $\mathbf{d}_j$,
${}^I \mathbf{R}_{L}$ and ${}^I \mathbf{p}_{L}$ are known IMU--LiDAR rotation and translation transformations, respectively.
${ }^G \mathbf{q}_j$ is a point on the nearest plane.
By substituting \eqref{eq:coupling} into \eqref{eq:residual} and solving equations,
Jacobians $\mathbf{H}_{k,j}$ and $\mathbf{H}_{k\!-\!N,j}$ with respect to $k$-th and $(k\!-\!N)\text{-th}$ states can be calculated as follows:
\begin{align}
    \mathbf{H}_{k,j} &=\left.\frac{\partial \mathbf{h}_j\left(\widehat{\mathbf{x}}_k \boxplus \widetilde{\mathbf{x}}_k,\,{ }^{L}\mathbf{n}_j\right)}{\partial \widetilde{\mathbf{x}}_k}\right|_{\widetilde{\mathbf{x}}_k=0} \nonumber\\
    &= \normalvector_j^{\top} \! \left[\,\Theta_{0,0} \ \, \mathbf{I}_{3} \ \, \mathbf{0}_{3\times V_0}\,\right], \nonumber\\
    \mathbf{H}_{k\!-\!N,j} &= \normalvector_j^{\top} \! \left[\,\Theta_{0,N} \ \, \mathbf{I}_{3} \ \cdots \ \Theta_{N,N} \ \, \mathbf{I}_{3} \ \, \mathbf{0}_{3\times V_N}\,\right], \label{eq:jacobian}\\
    \Theta_{m,r} & \triangleq -\widehat{\mathbf{R}}_{k\!-\!m}\!\left[\,\widehat{\mathbf{R}}_{k\!-\!m}^{\top}\widehat{\mathbf{R}}_{k\!-\!r}\mathbf{D}_j\,\right]_{\times}\!, \nonumber
\end{align}

\noindent where $V_m\!=\!12+6(s_w\!-\!m)$, $0 \! \leq \! m \! \leq \! r \! \leq \! N \! \leq \! s_w$, $m,r,N \! \in \! \mathbb{Z}^{0+}\!$,
$[\cdot]_{\times}$ denotes the skew-symmetric operator, and $\mathbf{I}_{3}$ is the $3\times3$ identity.
Note that Jacobians of past residuals with respect to the current state are not zero.
Hence, past states and their associated measurements can influence the current state estimation. 
In addition, augmented measurements mitigate the sparsity and thus reduce the risk of Jacobian singularity.

\textbf{Schmidt-Kalman update:}
Unlike the standard Kalman update, the Schmidt-Kalman update~\cite{schmidt1966acs-schmidt} is used to partially optimize the current and active states while preserving the fixed states.
During the iterative state update in \eqref{eq:iteration}, Kalman gain of fixed states $\mathbf{K}_f$ is set to zero
to treat the fixed states as nuisance parameters and not to update them, as follows:
\begin{equation}\label{eq:kalman-gain}
    \mathbf{K}=\left[\begin{array}{c}
        \mathbf{K}_u \\
        \mathbf{K}_f
        \end{array}\right]
        =
        \left[\begin{array}{c}
        \mathbf{K}_u \\
        \mathbf{0}
        \end{array}\right],
\end{equation}

\noindent where subscript $f$ and $u$ represent the fixed and updating (current and active) states, respectively.
In \eqref{eq:state-update}, the covariance update is written in the Joseph form to ensure numerical stability and preserve symmetry~\cite{joseph1968filter-joseph,schmidt1966acs-schmidt}, as the Kalman gain is deliberately adjusted.
Substituting \eqref{eq:kalman-gain} into \eqref{eq:state-update} yields:
\begin{equation}\label{eq:joseph-form}
    \begin{aligned}
        \mathbf{P} &= 
        {\setlength{\arraycolsep}{3pt}
        \left[
            \begin{array}{cc}
                \mathbf{P}_{uu} & \mathbf{P}_{uf} \\
                \mathbf{P}_{fu} & \mathbf{P}_{ff}
            \end{array}
        \right] }, \\
        \overline{\mathbf{P}}_k &= \mathbf{P} -
        {\setlength{\arraycolsep}{3.1pt}
        \left[
            \begin{array}{cc}
                \mathbf{K}_u (\mathbf{H} \mathbf{P} \mathbf{H}^{\transpose} + \mathbf{C} ) \mathbf{K}_u^{\transpose} & 
                \mathbf{K}_u \mathbf{H} 
                \left[
                    \begin{array}{c}
                        \mathbf{P}_{uu} \\
                        \mathbf{P}_{fu}
                    \end{array}
                \right] \\
                \left[
                    \begin{array}{c}
                        \mathbf{P}_{uu} \\
                        \mathbf{P}_{fu}
                    \end{array}
                \right]^{\transpose}
                \mathbf{H}^{\transpose} \mathbf{K}_u^{\transpose} & 
                \mathbf{0}
            \end{array}
        \right]. }
    \end{aligned}
\end{equation}

\noindent The covariances of fixed states remain unchanged but influence the covariance update of updating states,
affecting the calculation of Kalman gain in \eqref{eq:iteration}.
Consequently, fixed states guide the estimation of current and active states.

Unlike the classic Schmidt-Kalman filter~\cite{schmidt1966acs-schmidt} which treats sensor biases as fixed nuisance parameters,
our method applies this concept to poses.
While PUMA-LIO~\cite{jiang2022iros-pumalio} also fixes them, it fixes all past states without accounting for degeneracy,
which may overly constrain the estimation.
In contrast, we adaptively classify and slide states as shown in the following subsection.

\textbf{Degeneracy-aware sliding modes:}
As fixed states constrain the estimation of current and active states, only non-degenerate states should be kept in the fixed states set.
As discussed in \secref{sec:related-works}, we adopt the condition number as a practical degeneracy indicator,
as it captures both measurement sparsity and imbalance.
The condition number is defined as follows:
\begin{equation}\label{eq:condition-number}
    \chi(\mathbf{H}) \triangleq \frac{\sigma(\mathbf{H})_{\text{max}}}{\sigma(\mathbf{H})_{\text{min}}} \in \mathbb{R}^{+},\\
\end{equation}

\noindent where $\sigma(\mathbf{H})_{\text{max}}$ and $\sigma(\mathbf{H})_{\text{min}}$ are the maximum and minimum singular values of $\mathbf{H}$, respectively.
DA-ASKF uses two sliding modes (see \subfigref{fig:lodestar}{b}): full and partial.
At each filter update, current and active states slide by one frame,
and one sliding mode is chosen by the degeneracy level of the last active state $\mathbf{x}_{s_a}$.
If the state is non-degenerate $\left(\chi(\mathbf{H}(\mathbf{x}_{s_a})) \!<\! \mathcal{T}_\chi\right)$, it is transferred to the fixed state set (full sliding);
otherwise, it is removed from the sliding window (partial sliding).
$\mathcal{T}_{\chi}$ denotes the user-defined threshold of the condition number.

As a result, only non-degenerate states are retained in the fixed set, and degenerate states are removed from the window.
Moreover, active states are further optimized until the last slot in the window, enhancing the reliability of fixed states.

\textbf{Effect of DA-ASKF:}
By leveraging Schmidt-Kalman update and degeneracy-aware sliding modes,
DA-ASKF adaptively utilizes past states and measurements to constrain the estimation and mitigate measurement sparsity,
improving robustness and accuracy.
As noted in \secref{sec:related-works}, our stance is practical: stable under short-term degeneracy with non-zero measurements;
yet, prolonged zero-return remains challenging.

\subsection{Degeneracy-Aware Data Exploitation}\label{sec:da-de}
\textbf{Goal of DA-DE:}
To accurately optimize the current and active states, DA-DE selectively exploits measurements with high localizability contribution.
Furthermore, it reduces the risk of singularity in $\mathbf{H}$ by compensating for imbalance using informative measurements from fixed states.
The flow of DA-DE is illustrated in \subfigref{fig:lodestar}{c}.

\textbf{Localizability contribution:}
Since singular value decomposition (SVD) of $\mathbf{H}$ is equivalent to eigenvalue decomposition of the Hessian matrix $\mathbf{H}^{\transpose} \mathbf{H}$,
we can separate $\mathbf{H}$ into rotation and position components as follows:
\begin{equation}\label{eq:hessian}
    \renewcommand{\arraystretch}{1.1} 
    \mathbf{H}^{\transpose} \mathbf{H} = \left[\begin{array}{cc}
        {\mathbf{H}}_{\mathbf{R}}^{\transpose} \mathbf{H}_{\mathbf{R}} & {\mathbf{H}}_{\mathbf{R}}^{\transpose} \mathbf{H}_{\mathbf{p}} \\
        {\mathbf{H}}_{\mathbf{p}}^{\transpose} \mathbf{H}_{\mathbf{R}} & {\mathbf{H}}_{\mathbf{p}}^{\transpose} \mathbf{H}_{\mathbf{p}}
        \end{array}\right],
\end{equation}

\noindent where $\mathbf{H}_{\mathbf{R}}$ and $\mathbf{H}_{\mathbf{p}}$ are Jacobians with respect to rotation and position, respectively.
Then, the SVDs of $\mathbf{H}_{\mathbf{R}}$ and $\mathbf{H}_{\mathbf{p}}$, and the localizability contribution of $\mathbf{d}_j$ are computed as follows:
\begin{equation}\label{eq:localizability}
	\begin{aligned}
        \mathbf{H}_{\mathbf{p}}&=\mathbf{U}_{\mathbf{p}}\mathbf{\Sigma}_{\mathbf{p}}\mathbf{V}_{\mathbf{p}}^{\transpose}, \ \
        \mathbf{H}_{\mathbf{R}}=\mathbf{U}_{\mathbf{R}}\mathbf{\Sigma}_{\mathbf{R}}\mathbf{V}_{\mathbf{R}}^{\transpose},\\
        \boldsymbol{\Omega}_{\mathbf{p},j} &= \normalvector_j^{\transpose} \mathbf{V}_{\mathbf{p}}  \in \mathbb{R}^{3},\\
        \boldsymbol{\Omega}_{\mathbf{R},j} &= - \normalvector_j^{\transpose} \widehat{\mathbf{R}}_{k}\left[\mathbf{D}_j\right]_{\times} \mathbf{V}_{\mathbf{R}}  \in \mathbb{R}^{3},
    \end{aligned}
\end{equation}

\noindent where $\mathbf{U}$, $\mathbf{V}$, and $\mathbf{\Sigma}$ are left and right singular vector matrices and descending ordered singular value matrix of $\mathbf{H}$, respectively.
The rotational and translational localizability contribution $\boldsymbol{\Omega}_{\mathbf{R},j}$ and $\boldsymbol{\Omega}_{\mathbf{p},j}$ of each measurement
are computed by projecting the rotational and translational components
of $\mathbf{H}_j$ (see \eqref{eq:jacobian}) onto $\mathbf{V}_{\mathbf{R}}$ and $\mathbf{V}_{\mathbf{p}}$, respectively, similar to~\cite{tuna2023tro-xicp}.

\textbf{Data exploitation:}
The process consists of three steps:

\noindent (\romsmall{1}) \textit{Pruning:} 
Measurements with low localizability contribution cannot be easily distinguished from the noise and the outliers~\cite{tuna2023tro-xicp}.
Thus, only measurements with high contribution are retained for the current and active states, as follows:
\begin{equation}\label{eq:pruning}
    \mathbf{\Psi}_u = \left\{ \mathbf{d}_{j} \,\middle|\, \underset{{\ell \in 1,2,3}}{\max} \left( \max\left( \boldsymbol{\Omega}_{\mathbf{p},j}^{\ell}, \, \boldsymbol{\Omega}_{\mathbf{R},j}^{\ell} \right) \right) > \mathcal{T}_{\text{loc}} \right\},
\end{equation}

\noindent where $\mathcal{T}_{\text{loc}}$ is the user-defined threshold for the localizability contribution
so that only measurements with strong geometric alignment are retained.

\noindent (\romsmall{2}) \textit{SVD for condition number evaluation:}
The Jacobians from pruned data $\mathbf{\Psi}_u$ are stacked into $\overline{\mathbf{H}}$,
and SVD is applied to compute the updated condition number $\chi(\overline{\mathbf{H}})$ as \eqref{eq:condition-number}.
$\chi(\overline{\mathbf{H}})$ reveals the degenerate directions of current and active states,  
corresponding to the singular vectors associated with the smallest singular values due to measurement imbalance.

\noindent (\romsmall{3}) \textit{Compensating degenerate directions:}
Measurements from fixed states are sorted by their contribution along the degenerate directions
$\overline{\mathbf{V}}_{\mathbf{p}}^3$, $\overline{\mathbf{V}}_{\mathbf{R}}^3$, 
and incrementally added in descending 

\noindent order until $\chi(\overline{\mathbf{H}})$ falls below $\mathcal{T}_{\chi}$, as follows:
\begin{equation}\label{eq:compensation}
    \begin{aligned}
        \boldsymbol{\Omega}_{\mathbf{p},j}^3 &= \normalvector_j^{\transpose} \overline{\mathbf{V}}_{\mathbf{p}}^3  \in \mathbb{R}^{+},\\
        \boldsymbol{\Omega}_{\mathbf{R},j}^3 &= - \normalvector_j^{\transpose} \widehat{\mathbf{R}}_{k}\left[\mathbf{D}_j\right]_{\times} \overline{\mathbf{V}}_{\mathbf{R}}^3  \in \mathbb{R}^{+},\\
        \mathbf{\Psi}_f &\leftarrow \mathbf{\Psi}_f \bigcup \left\{ \mathbf{d}_{j} \,\middle|\, \max\left( \boldsymbol{\Omega}_{\mathbf{p},j}^3, \, \boldsymbol{\Omega}_{\mathbf{R},j}^3 \right) > \mathcal{T}_{\text{loc}} \right\}.
    \end{aligned}
\end{equation}

\noindent The final set of exploited data used in the filter update is obtained as $\mathbf{\Psi}_\text{all} = \mathbf{\Psi}_f \bigcup \mathbf{\Psi}_u$.

\textbf{Effect of DA-DE:} DA-DE only uses informative measurements, precisely identifies degenerate directions of states,
and compensates for imbalance using measurements from reliable fixed states.
This prevents the fixed states from over-constraining the estimation,
reduces the condition number, and improves numerical stability and robustness.

\section{Experimental Results}\label{sec:experiments}
\subsection{Experimental Setup}\label{sec:setup}
To evaluate the performance of LODESTAR, we conduct extensive experiments on five public datasets:
\dataset{2021\;HILTI}~\cite{helmberger2022ral-2021hilti}, \dataset{2022\;HILTI}~\cite{zhang2022ral-2022hilti}, \dataset{NTU-VIRAL}~\cite{nguyen2022ijrr-ntuviral}, \dataset{Newer\;College}~\cite{ramezani2020iros-ncd}, and \dataset{SubT-MRS}~\cite{zhao2024cvpr-subtmrs}.
For \dataset{2021\;HILTI}, we use sequences from both confined indoor (\dataset{Basement\,1}, \dataset{Basement\,4}, \dataset{Drone\;Arena})
and open outdoor (\dataset{Campus\,2}, \dataset{Construction\,2}).
For \dataset{2022\;HILTI}, the \dataset{Long\;Corridor} sequence is selected.
For \dataset{NTU-VIRAL}, we use \dataset{SPMS} sequences, which are collected in high-altitude outdoor environments.
For \dataset{SubT-MRS}, \dataset{Long\;Corridor} and \dataset{Cavern\;Handheld} sequences are selected, which exhibit highly imbalanced and unstructured environments, respectively.
The \dataset{Newer\;College} dataset is used as a representative of non-degenerate environments, collected in open campus areas.

Note that only one LiDAR is used in multi-LiDAR datasets:
Livox Mid-70 for \dataset{2021\;HILTI} and horizontal OS1-16 for \dataset{NTU-VIRAL}.
Similarly, other datasets utilize a single LiDAR:
VLP-16 for \dataset{SubT-MRS}, PandarXT-32 for \dataset{2022\;HILTI}, and OS1-64 for \dataset{Newer\;College}.
Consequently, all datasets except \dataset{Newer\;College} exhibit substantial measurement imbalance and sparsity,
due to both the small number of LiDAR channels and geometric degeneracy in the environments.

To validate the contribution of the proposed DA-ASKF and DA-DE,
we perform an ablation study by comparing with existing degeneracy-aware modules integrated with a common baseline (FAST-LIO2)~\cite{xu2022tro-fastlio2}:  
NA-LOAM~\cite{yang2024sensors-naloam} (imbalance-aware weight adjustment);
GenZ-ICP~\cite{lee2024ral-genzicp} and AdaLIO~\cite{lim2023ur-adalio} (sparsity-aware data augmentation);
and the method by Hinduja~\etal~\cite{hinduja2019iros-hinduja} (degeneracy-aware constrained optimization).
We also compare with two naive implementations: vanilla sliding window (Vanilla SW) that updates all states;
and vanilla Schmidt-Kalman filter (Vanilla Schmidt) that fixes all past states.
Both methods use all measurements without selection.

Additionally, for comparison with state-of-the-art methods,
we consider both LiDAR-only odometry (LO): KISS-ICP~\cite{vizzo2023ral-kissicp}, GenZ-ICP~\cite{lee2024ral-genzicp}, DLO~\cite{chen2022ral-dlo};
and LiDAR-inertial odometry (LIO): LIO-SAM~\cite{shan2020iros-liosam}, DLIO~\cite{chen2023icra-dlio}, iG-LIO~\cite{chen2024ral-iglio}, AdaLIO~\cite{lim2023ur-adalio}.

The absolute pose error (APE) with EVO~\cite{grupp2017evo} is used as the main metric to effectively capture the accumulated drift from degeneracy.
For \dataset{2022\;HILTI} and \dataset{NTU-VIRAL}, official evaluation tools are used~\cite{helmberger2022ral-2021hilti,nguyen2022ijrr-ntuviral}.
Mapping results are qualitatively compared.
The voxel resolution is set to 0.3\,m for degenerate datasets (\dataset{2021\;HILTI}, \dataset{2022\;HILTI}, \dataset{NTU-VIRAL}, and \dataset{SubT-MRS})
and 0.5\,m for the non-degenerate \dataset{Newer\;College} across all methods to ensure fair comparisons, given the sensitivity of LIO/LO methods to voxel size.
Accordingly, results may differ from the original papers.
Other parameters follow publicly released default values of each method (no sequence-specific tuning).
All configurations and self-implemented degeneracy-aware modules are included in the supplementary material.
We empirically set $s_a\!=\!2$ and $s_f\!=\!2$ for all datasets.
\subsection{Ablation Study: Effectiveness of DA-ASKF and DA-DE}\label{sec:ablation}
\textbf{Threshold sensitivity:}
To examine the sensitivity of DA-ASKF and DA-DE to user-defined thresholds, we performed a small grid search on two imbalance or sparsity-dominant sequences
with all other settings fixed (\tabref{tab:threshold_ablation}).
The APE is minimized at $\mathcal{T}_{\chi}\!=\!1.5$ and $\mathcal{T}_{\text{loc}}\!=\!\cos 35^\circ$.
As $\mathcal{T}_{\chi}\!=\!1$ implies ideal isotropy, a value of $1.5$ serves as a practical threshold--neither universal nor optimal--conservatively storing states as fixed.
Similarly, deviating $\mathcal{T}_{\text{loc}}$ either admits noisy, less-informative measurements or induces sparsity by over-pruning;
$\cos 35^\circ$ is a stable middle ground. We fix $\mathcal{T}_{\chi}\!=\!1.5$ and $\mathcal{T}_{\text{loc}}\!=\!\cos 35^\circ$ for all datasets.

\textbf{Accuracy and robustness:}
The quantitative results are reported in \tabref{tab:ape_ablation} and \tabref{tab:score_ablation}.
The NA-LOAM module did not show meaningful improvements and diverged in \dataset{Cavern\;Handheld},
as it arbitrarily adjusts the weights of measurements without considering their contribution.
Although AdaLIO increases the number of points, it does not consider measurement imbalance and diverged in \dataset{Cavern\;Handheld}.
GenZ-ICP combines point-to-plane with additional point-to-point registration,
which is known to be susceptible to noise and outliers~\cite{rusinkiewicz2001point-to-plane}, and therefore showed divergence and drift.  
The method by Hinduja~\etal showed improvements only in \dataset{Basement\,4}, as it simply avoids updating degenerate states.  
Vanilla SW showed slight improvement in \dataset{Cavern\;Handheld} and \dataset{Basement\,4}, due to increased data.
Similarly, Vanilla Schmidt showed improved accuracy in \dataset{Basement\,4}.
However, these methods over-constrain the estimation, resulting in degraded performance in the \dataset{Long\;Corridor} sequences.
In particular, Vanilla Schmidt treats all past states as fixed references, which caused divergence in \dataset{Cavern\;Handheld}.

\begin{table}[t]
    \centering
    \caption{Absolute Pose Errors with Different Thresholds}\label{tab:threshold_ablation}
    \vstabcap
    \scriptsize
    \renewcommand{\arraystretch}{1.07}
    \setlength\tabcolsep{4.5pt} 
    \begingroup 
    \setlength{\aboverulesep}{0pt} 
    \setlength{\belowrulesep}{0.2pt} 
    \begin{tabular}{lcccccc}
        \thline
        & \multicolumn{3}{c}{\datasettable{SubT-MRS-Long\;Corridor}} & \multicolumn{3}{c}{\datasettable{2021\;HILTI-Basement\,4}} \\ \cmidrule[0.4pt](lr){2-4}\cmidrule[0.4pt](lr){5-7}
        & \multicolumn{3}{c}{$\mathcal{T}_{\text{loc}}$} & \multicolumn{3}{c}{$\mathcal{T}_{\text{loc}}$} \\
        $\mathcal{T}_{\chi}$ & \(\cos30^\circ\) & \(\cos35^\circ\) & \(\cos45^\circ\) & \(\cos30^\circ\) & \(\cos35^\circ\) & \(\cos45^\circ\) \\ \cmidrule[0.4pt](lr){1-4}\cmidrule[0.4pt](lr){5-7}
        1.5 & 1.551 & \textbf{1.386} & 1.775 & 0.189 & \textbf{0.089} & 0.120 \\
        2.0 & 2.305 & 2.291 & 1.660 & 0.246 & 0.278 & 0.276 \\
        3.0 & 2.486 & 1.595 & 2.259 & 0.179 & 0.124 & 0.293 \\
        \thline
    \end{tabular}
    \endgroup
    {\scriptsize \parbox{0.84\linewidth}{\vstabfoot The best result is shown in \textbf{bold}. Units are in meters.}}
    \vspace{-0.24cm}
\end{table}

\begin{figure}[t]
    \centering
    \includegraphics[width=0.46\textwidth]{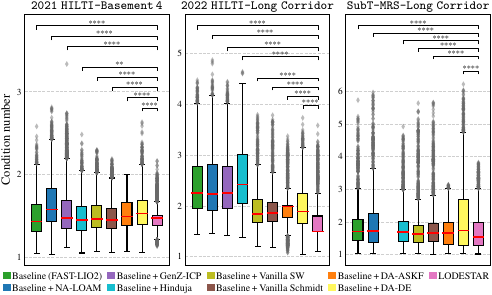}
    \\ \vsfigcap
    \caption{Condition number comparison of different degeneracy-aware modules.
    Baseline\,+\,GenZ-ICP diverged in \dataset{SubT-MRS}--\dataset{Long\;Corridor}, and is thus excluded.
    The **** and ** denote that \textit{p}-values of the paired \textit{t}-test are less than $10^{-4}$ and $10^{-2}$, respectively, indicating statistically significant differences.}
    \label{fig:boxplot}
    \vspace{-0.22cm}
\end{figure}

\begin{table}[t]
    \centering
    \caption{Absolute Pose Errors from Different Degeneracy-Aware Modules Across Three Sequences.}\label{tab:ape_ablation}
    \vstabcap
    \scriptsize
    \renewcommand{\arraystretch}{1.07}
    \setlength\tabcolsep{4.7pt} 
    \begingroup 
    \setlength{\aboverulesep}{0pt} 
    \setlength{\belowrulesep}{0pt} 
    \begin{tabular}{lccc}
        \thline
        & \multicolumn{2}{c}{\datasettable{SubT-MRS}}    & \datasettable{2021\;HILTI} \\
        \cmidrule[0.4pt](lr){2-3} \cmidrule[0.4pt](lr){4-4}
        \multicolumn{1}{c}{Method} & \datasettable{Long\;Corridor} & \datasettable{Cavern\;Handheld} & \datasettable{Basement\,4} \\
        \hline
        Baseline (FAST-LIO2)~\cite{xu2022tro-fastlio2}   & 3.113        & 3.521          & 0.503     \\
        Baseline\,+\,NA-LOAM~\cite{yang2024sensors-naloam} & 3.325        & $\times$              & 0.497     \\
        Baseline\,+\,GenZ-ICP~\cite{lee2024ral-genzicp} & $\times$            & $\times$              & 6.365     \\
        Baseline\,+\,AdaLIO~\cite{lim2023ur-adalio} & 3.160        & $\times$              & 0.492     \\
        Baseline\,+\,Hinduja~\cite{hinduja2019iros-hinduja} & 3.674        & 4.493          & 0.385     \\
        Baseline\,+\,Vanilla SW & 3.962        & \hl{ 1.460 }          & 0.405     \\
        Baseline\,+\,Vanilla Schmidt & 4.283        & $\times$              & 0.401     \\
        Baseline\,+\,DA-ASKF & \hl{ 1.833 }        & 1.774          & \hl{ 0.192 }     \\
        Baseline\,+\,DA-DE & 2.082        & 3.899          & 0.332     \\
        \makecell{LODESTAR}                         & \textbf{1.386}        & \textbf{0.671}          & \textbf{0.089}  \\
        \thline
    \end{tabular}
    \endgroup 
    {\scriptsize \parbox{0.94\linewidth}{\vstabfoot The best and second-best results are shown in \textbf{bold} and shaded \hl{gray}, respectively. ``$\times$'' indicates the divergence. Units are in meters.}}
    \vspace{-0.2cm}
\end{table}

\begin{table}[t]
    \centering
    \caption{Scores from Different Degeneracy-Aware Modules on The \dataset{2022\;HILTI}-\dataset{Long\;Corridor} Sequence.}\label{tab:score_ablation}
    \vstabcap
    \scriptsize
    \renewcommand{\arraystretch}{1.07}
    \setlength\tabcolsep{2.8pt} 
    \begin{tabular}{lcccccc}
        \thline
        \multicolumn{1}{c}{Method} & \textless{}1\,cm & \textless{}3\,cm & \textless{}6\,cm & \textless{}10\,cm & \textgreater{}10\,cm & Score $\uparrow$ \\
        \hline
        Baseline (FAST-LIO2)~\cite{xu2022tro-fastlio2}            & 0              & 1              & 3              & 2               & 0                  & \hl{ 28.33 } \\
        Baseline\,+\,NA-LOAM~\cite{yang2024sensors-naloam}& 0              & 1              & 2              & 3               & 0                  & 25.00 \\
        Baseline\,+\,GenZ-ICP~\cite{lee2024ral-genzicp}& 0              & 1              & 2              & 2               & 1                  & 23.33 \\
        Baseline\,+\,AdaLIO~\cite{lim2023ur-adalio}& 0              & 1              & 2              & 3               & 0                  & 25.00 \\
        Baseline\,+\,Hinduja~\cite{hinduja2019iros-hinduja}& 0              & 0              & 1              & 3               & 2                  & 10.00 \\
        Baseline\,+\,Vanilla SW    & 0              & 0              & 0              & 0               & 6                  & 0.00  \\
        Baseline\,+\,Vanilla Schmidt           & 0              & 0              & 0              & 0               & 6                  & 0.00  \\
        Baseline\,+\,DA-ASKF  & 0              & 1              & 1              & 1               & 3                  & 16.67 \\
        Baseline\,+\,DA-DE & 0              & 1              & 0              & 2               & 3                  & 13.33 \\
        \makecell{LODESTAR}                         & 0              & 2              & 2              & 2               & 0                  & \textbf{33.33} \\
        \thline
    \end{tabular}
    {\scriptsize \parbox{0.95\linewidth}{\vstabfoot The best and second-best results are shown in \textbf{bold} and shaded \hl{gray}, respectively.}}
    \vspace{-0.22cm}
\end{table}

In contrast, the proposed DA-ASKF and DA-DE modules effectively mitigate measurement sparsity and imbalance, respectively,
yielding improved accuracy across all sequences in \tabref{tab:ape_ablation}, except for DA-DE in \dataset{Cavern\;Handheld}, which is discussed further below.
For \dataset{2022\;HILTI} (see \tabref{tab:score_ablation}), applying DA-ASKF alone over-constrained the estimation due to abundant measurements from the densely scanning LiDAR.
Conversely, DA-DE alone was unable to fully address measurement imbalance, leading to suboptimal performance.

These results highlight the importance of combining both modules to achieve balanced and adaptive constraints.
DA-ASKF improves accuracy by leveraging adaptive Schmidt-Kalman update and fixing non-degenerate states,
which serve as reference anchors that reliably constrain current and active states.  
DA-DE enhances robustness by selecting informative measurements, improving both balance and numerical stability.
When combined, DA-ASKF and DA-DE enable LODESTAR to simultaneously handle measurement sparsity, imbalance,  
and degeneracy-aware constrained optimization, thereby achieving the best overall performance.

\textbf{Condition number analysis:}
We use the condition number as a practical degeneracy indicator--not a certificate of optimal solvability--as it reflects measurement sparsity, imbalance,
and the risk of Jacobian singularity~\cite{tuna2025tfr,yang2024sensors-naloam,lee2024ral-genzicp,cheney1998condition-number,zhang2016icra-zhang,hinduja2019iros-hinduja,tuna2023tro-xicp}.
Across all sequences in \figref{fig:boxplot}, LODESTAR yielded competitive maxima, medians, and interquartile ranges (IQRs):
the narrowest IQR in \dataset{Basement\,4}; the lowest medians in two \dataset{Long\;Corridor} sequences;
and the lowest maxima among all modules except Baseline$\,+\,$DA-ASKF.
Using DA-ASKF alone densifies measurements, thereby lowering maxima and narrowing the overall spread;
however, its medians remain higher than LODESTAR's, indicating unresolved imbalance.
Using DA-DE alone enlarges the spread and maxima because aggressive selection increases measurement sparsity without fully resolving imbalance.
By combining DA-ASKF (mitigating sparsity) and DA-DE (reducing imbalance),
LODESTAR achieves consistently lower medians with small spreads.
While low condition numbers do not optimally guarantee superior accuracy,
the lower medians of LODESTAR align with the improved localization metrics in \tabref{tab:ape_ablation} and \tabref{tab:score_ablation},
supporting the condition number's role as a practical indicator.


\begin{table*}[t]
    \centering
    \caption{Absolute Pose Errors and Average Computation Times of LODESTAR and State-of-the-Art Methods.}\label{tab:ape_sota}
    \vstabcap
    \scriptsize
    \renewcommand{\arraystretch}{1.065}
    \setlength\tabcolsep{4.5pt} 
    \begingroup 
    \setlength{\aboverulesep}{0pt} 
    \setlength{\belowrulesep}{0pt} 
    \begin{tabular}{ccccccccccc}
        \thline
        & & \multicolumn{3}{c}{LO} & \multicolumn{6}{c}{LIO} \\
        \cmidrule[0.4pt](lr){3-5} \cmidrule[0.4pt](lr){6-11}
        Dataset & Sequence & KISS-ICP~\cite{vizzo2023ral-kissicp} & GenZ-ICP~\cite{lee2024ral-genzicp} & DLO~\cite{chen2022ral-dlo} & LIO-SAM~\cite{shan2020iros-liosam} & DLIO~\cite{chen2023icra-dlio} & iG-LIO~\cite{chen2024ral-iglio} & AdaLIO~\cite{lim2023ur-adalio} & Baseline~\cite{xu2022tro-fastlio2} & LODESTAR \\ \hline
        \multirow{2}{*}{\makecell{\datasettable{SubT-}\\\datasettable{MRS}}} & \datasettable{Long\;Corridor} & 3.028 & \hl{ 1.982 } & 9.416 & 2.959 & 2.294 & 3.136 & 3.160 & 3.113 & \textbf{1.386} \\
        & \datasettable{Cavern\;Handheld} & 9.525 & 2.938 & 3.891 & $\times$ & \textbf{0.585} & 0.710 & $\times$ & 3.521 & \hl{ 0.671 } \\
        \hline
        \multirow{5}{*}{\makecell{\datasettable{2021}\\\datasettable{HILTI}}} & \datasettable{Basement\,1} & 7.718 & 5.055 & 7.811 & - & 3.052 & 0.197 & \textbf{0.168} & 0.318 & \hl{ 0.174 } \\
        & \datasettable{Basement\,4} & 2.337 & 1.993 & 1.880 & - & 2.123 & 0.556 & \hl{ 0.492 } & 0.503 & \textbf{0.089} \\
        & \datasettable{Campus\,2} & 18.058 & 14.065 & 15.279 & - & 0.228 & 0.082 & 0.060 & \hl{ 0.058 } & \textbf{0.052} \\
        & \datasettable{Construction\,2} & 7.773 & $\times$ & 21.799 & - & 0.779 & \textbf{0.064} & 0.167 & 0.165 & \hl{ 0.078 } \\
        & \datasettable{Drone\;Arena} & 5.882 & 8.575 & 7.621 & - & $\times$ & 0.221 & \hl{ 0.211 } & \hl{ 0.211 } & \textbf{0.206} \\
        \hline
        \multirow{3}{*}{\makecell{\datasettable{NTU-}\\\datasettable{VIRAL}}} & \datasettable{SPMS\,01} & $\times$ & $\times$ & 26.430 & 1.296 & \hl{ 1.105 } & \textbf{0.237} & 1.680 & 2.002 & 1.582 \\
        & \datasettable{SPMS\,02} & $\times$ & $\times$ & 28.119 & $\times$ & \hl{ 1.171 } & $\times$ & 2.863 & 2.973 & \textbf{0.610} \\
        & \datasettable{SPMS\,03} & $\times$ & $\times$ & 24.087 & $\times$ & 2.062 & $\times$ & \hl{ 0.833 } & 0.839 & \textbf{0.582} \\
        \hline
        \multirow{3}{*}{\makecell{\datasettable{Newer}\\\datasettable{College}}} & \datasettable{Short\;exp} & 0.744 & 0.635 & \textbf{0.368} & 3.973 & 0.460 & \hl{ 0.369 } & 0.430 & 0.412 & 0.429 \\
        & \datasettable{Long\;exp} & 8.305 & $\times$ & 0.408 & 7.621 & 0.447 & 0.428 & 0.404 & \hl{ 0.392 } & \textbf{0.385} \\
        & \datasettable{Quad\;dynamics} & $\times$ & 0.125 & 0.130 & 0.153 & 0.174 & \hl{ 0.122 } & 0.129 & 0.125 & \textbf{0.118} \\
        \hline
        \hline
        \multicolumn{2}{c}{Average computation time} & 46.49 & 22.31 & \hl{ 13.16 } & 33.09 & 44.85 & \textbf{9.96} & 19.12 & 21.04 & 28.69 \\
        \thline
    \end{tabular}
    \endgroup 
    {\scriptsize \parbox{0.95\linewidth}{\vstabfoot The best and second-best results are shown in \textbf{bold} and shaded \hl{gray}, respectively. ``$\times$'' indicates the divergence. ``-'' denotes that the algorithm does not support the given sensor. Units for pose error and computation time are meters and milliseconds, respectively.}}
    \vspace{-0.33cm}
\end{table*}
\begin{figure*}[t]
    \centering
    \includegraphics[width=0.93\textwidth]{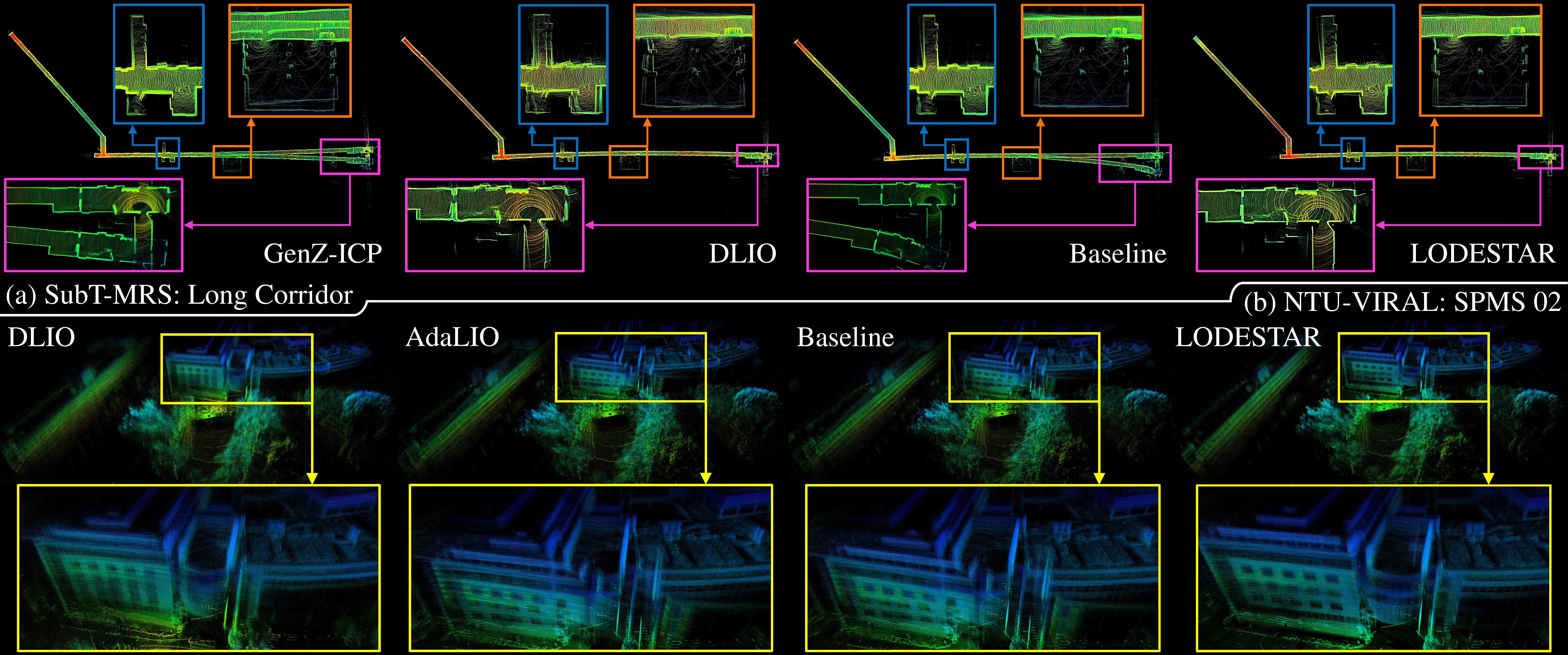}
    \\ \vsfigcap
    \caption{Mapping results of LODESTAR and the top three state-of-the-art methods on  
    (a) \dataset{SubT-MRS}--\dataset{Long\;Corridor} and (b) \dataset{NTU-VIRAL}--\dataset{SPMS\,02} sequences.  
    Despite imbalanced or sparse measurements,
    LODESTAR consistently mapped the environment with minimal drift, whereas others exhibited noticeable drift.}
    \label{fig:mapping}
    \vspace{-0.32cm}
\end{figure*}


\begin{table}[t]
    \vspace{-0.14cm}
    \centering
    \caption{Scores from LODESTAR and State-of-the-Art Methods on The \dataset{2022\;HILTI}-\dataset{Long\;Corridor} Sequence.}\label{tab:score_sota}
    \vstabcap
    \scriptsize
    \renewcommand{\arraystretch}{1.06}
    \setlength\tabcolsep{3.6pt} 
    \begin{tabular}{cccccccc}
        \thline
        Type & Method & \textless{}1\,cm & \textless{}3\,cm & \textless{}6\,cm & \textless{}10\,cm & \textgreater{}10\,cm & Score $\uparrow$ \\
        \hline
        \multirow{3}{*}{LO}&KISS-ICP~\cite{vizzo2023ral-kissicp}& 0  & 0  & 0  & 0   & 6      & 0.00 \\
        &GenZ-ICP~\cite{lee2024ral-genzicp}& 0  & 0  & 0  & 0   & 6  & 0.00 \\
        &DLO~\cite{chen2022ral-dlo}& 0  & 1  & 0  & 2   & 3      & 13.33 \\
        \hline
        \multirow{6}{*}{LIO}&LIO-SAM~\cite{shan2020iros-liosam}    & 0  & 0  & 0  & 0   & 6      & 0.00  \\
        &DLIO~\cite{chen2023icra-dlio}   & 0  & 1  & 2  & 1   & 2      & 21.67  \\
        &iG-LIO~\cite{chen2024ral-iglio} & 0  & 1  & 5  & 0   & 0      & \textbf{35.00} \\
        &AdaLIO~\cite{lim2023ur-adalio}  & 0  & 1  & 2  & 3   & 0      & 25.00 \\
        &Baseline~\cite{xu2022tro-fastlio2}& 0  & 1  & 3  & 2   & 0      & 28.33 \\
        &LODESTAR & 0  & 2  & 2  & 2   & 0      & \hl{ 33.33 } \\
        \thline
    \end{tabular}
    {\scriptsize \parbox{0.94\linewidth}{\vstabfoot The best and second-best results are shown in \textbf{bold} and shaded \hl{gray}, respectively.}}
    \vspace{-0.4cm}
\end{table}

\subsection{Comparison with State-of-the-Art Methods}\label{sec:sota}
To further validate the performance of LODESTAR, we compared it with state-of-the-art methods.
The quantitative results are reported in \tabref{tab:ape_sota} and \tabref{tab:score_sota},
and the qualitative mapping results are shown in \figref{fig:mapping}.
Among LO methods, KISS-ICP struggled under degeneracy and failed in \dataset{SPMS} sequences.
GenZ-ICP achieved notable result in \dataset{SubT-MRS}-\dataset{Long\;Corridor}, as it explicitly targets corridors,
but does not generalize well to sparse or dynamic datasets like \dataset{2021\;HILTI} and \dataset{NTU-VIRAL}.
DLO is the only LO method that did not diverge across datasets,
though it is mainly effective in dense datasets like \dataset{2022\;HILTI} and \dataset{Newer\;College}.
LIO methods generally showed higher robustness except for LIO-SAM, which diverged or drifted in most datasets.
DLIO achieved competitive accuracy across most datasets.
iG-LIO performed comparably but diverged in sparse and dynamic cases as it mainly focuses on faster computation.
AdaLIO showed enhanced robustness to sparsity but failed in \dataset{Cavern\;Handheld}.

In contrast, LODESTAR achieved high accuracy across the degenerate datasets
by jointly addressing sparsity, imbalance, and constrained optimization through DA-ASKF and DA-DE.
It also demonstrated comparable performance in the non-degenerate dataset.
Qualitative mapping results showed improved consistency under degenerate conditions.

\subsection{Computation Time Analysis}\label{sec:time}
\begin{figure}[t]
    \centering
    \includegraphics[width=0.415\textwidth]{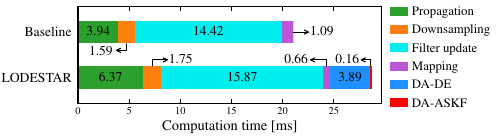}
    \\ \vsfigcap
    \caption{Average computation time of LODESTAR and the baseline~\cite{xu2022tro-fastlio2}.
    Overall, the computation time remains comparable to the baseline.}
    \label{fig:runtime}
    \vspace{-0.4cm}
\end{figure}

The computation time of LODESTAR compared with baseline and state-of-the-art methods is summarized in \tabref{tab:ape_sota} and visualized in \figref{fig:runtime}.
All tests were run on an Intel Core i9-13900K CPU with 32\,GB RAM.
LODESTAR achieved real-time performance on par with other methods,
while introducing minimal overhead compared with the baseline.
Slight increases were observed in filter propagation and update times, due to additional matrix operations from the increased number of states.
However, DA-DE reduces the number of points used, which shortens the mapping time and mitigates further increases in the filter update.
DA-ASKF introduces negligible overhead, as it mainly involves lightweight data copying.

\section{Conclusion}\label{sec:conclusion}
In this letter, we proposed LODESTAR, a degeneracy-aware LIO framework with DA-ASKF and DA-DE modules.
By enabling adaptive constrained optimization and selective data exploitation,
LODESTAR effectively addresses measurement sparsity and imbalance in a unified manner.
Extensive experiments validated its effectiveness across diverse degenerate scenarios,
demonstrating improved or comparable accuracy and robustness compared with existing LiDAR-based odometry methods and degeneracy-aware modules.
While LODESTAR handles short-term geometric degeneracy, it fails when LiDAR returns are absent.
Future work will explore LiDAR-radar fusion that exploits Doppler velocity to handle zero or near-zero LiDAR returns and to enhance robustness in adverse weather conditions such as rain and fog.
Please refer to the supplementary material for additional results, visualizations, and experimental configurations.





\bibliography{URL-bib}

@STRING{ral  = {IEEE Robot. Automat. Lett.} }

@STRING{tro  = {IEEE Trans. Robot.} }

@STRING{tfr  = {IEEE Trans. Field Robot.} }

@STRING{ijrr = {Int. J. Robot. Res.} }

@STRING{iros   = {Proc. IEEE/RSJ Int. Conf. Intell. Robot. Syst. (IROS)} }

@STRING{icra   = {Proc. IEEE Int. Conf. Robot. Automat. (ICRA)} }

@STRING{ur     = {Proc. Int. Conf. Ubiquit. Robot. (UR)} }

@STRING{threedim = {Proc. Int. Conf. 3-D Digit. Img. Model. (3DIM)} }

@STRING{cvpr  = {Proc. IEEE/CVF Conf. Comput. Vis. Pattern Recognit. (CVPR)} }

@STRING{InformationFusion = {Inf. Fusion} }

@inproceedings{zhao2024cvpr-subtmrs,
  title={{SubT-MRS dataset: Pushing SLAM towards all-weather environments}},
  author={Zhao, Shibo and Gao, Yuanjun and Wu, Tianhao and Singh, Damanpreet and Jiang, Rushan and Sun, Haoxiang and Sarawata, Mansi and Qiu, Yuheng and Whittaker, Warren and Higgins, Ian and others},
  booktitle=cvpr,
  pages={22647--22657},
  year={2024}
}

@article{helmberger2022ral-2021hilti,
  title={{The HILTI SLAM challenge dataset}},
  author={Helmberger, Michael and Morin, Kristian and Berner, Beda and Kumar, Nitish and Cioffi, Giovanni and Scaramuzza, Davide},
  journal=ral,
  volume={7},
  number={3},
  pages={7518--7525},
  year={2022},
}

@article{zhang2022ral-2022hilti,
  title={{HILTI-Oxford Dataset: A millimeter-accurate benchmark for simultaneous localization and mapping}},
  author={Zhang, Lintong and Helmberger, Michael and Fu, Lanke Frank Tarimo and Wisth, David and Camurri, Marco and Scaramuzza, Davide and Fallon, Maurice},
  journal=ral,
  volume={8},
  number={1},
  pages={408--415},
  year={2022},
}

@article{nguyen2022ijrr-ntuviral,
  title={{NTU VIRAL: A visual-inertial-ranging-LiDAR dataset, from an aerial vehicle viewpoint}},
  author={Nguyen, Thien-Minh and Yuan, Shenghai and Cao, Muqing and Lyu, Yang and Nguyen, Thien Hoang and Xie, Lihua},
  journal=ijrr,
  volume={41},
  number={3},
  pages={270--280},
  year={2022},
}

@inproceedings{ramezani2020iros-ncd,
  title={{The Newer College Dataset: Handheld LiDAR, inertial and vision with ground truth}}, 
  author={Ramezani, Milad and Wang, Yiduo and Camurri, Marco and Wisth, David and Mattamala, Matias and Fallon, Maurice},
  booktitle=iros, 
  year={2020},
  pages={4353-4360},
}

@article{jeon2021ral-run,
  title={{Run your visual-inertial odometry on NVIDIA Jetson: Benchmark tests on a micro aerial vehicle}},
  author={Jeon, Jinwoo and Jung, Sungwook and Lee, Eungchang and Choi, Duckyu and Myung, Hyun},
  journal=ral,
  volume={6},
  number={3},
  pages={5332--5339},
  year={2021},
}

@article{lim2024ijrr-quatro++,
  title={{Quatro++: Robust global registration exploiting ground segmentation for loop closing in LiDAR SLAM}},
  author={Lim, Hyungtae and Kim, Beomsoo and Kim, Daebeom and Lee, Eungchang Mason and Myung, Hyun},
  journal=ijrr,
  volume={43},
  number={5},
  pages={685--715},
  year={2024},
}

@article{kim2022ral-step,
  title={{STEP: State estimator for legged robots using a preintegrated foot velocity factor}},
  author={Kim, Yeeun and Yu, Byeongho and Lee, Eungchang Mason and Kim, Joon-Ha and Park, Hae-Won and Myung, Hyun},
  journal=ral,
  volume={7},
  number={2},
  pages={4456--4463},
  year={2022},
}

@book{nocedal1999gn-lm-gradient,
  title={{Numerical Optimization}},
  author={Nocedal, Jorge and Wright, Stephen J},
  year={1999},
  publisher={Springer}
}

@article{hertzberg2013infofusion-manifold,
  title={{Integrating generic sensor fusion algorithms with sound state representations through encapsulation of manifolds}},
  author={Hertzberg, Christoph and Wagner, Ren{\'e} and Frese, Udo and Schr{\"o}der, Lutz},
  journal=InformationFusion,
  volume={14},
  number={1},
  pages={57--77},
  year={2013},
}

@misc{grupp2017evo,
  author       = {Michael Grupp},
  title        = {{EVO: Python package for the evaluation of odometry and SLAM}},
  howpublished = {{2017. [Online]. Available: \url{https://github.com/MichaelGrupp/evo}}},
  note         = {{[Accessed: Apr. 10, 2025]}}
}

@inproceedings{lee2021iros-real,
	title={{REAL: Rapid exploration with active loop-closing toward large-scale 3D mapping using UAVs}},
	author={Lee, Eungchang Mason and Choi, Junho and Lim, Hyungtae and Myung, Hyun},
	booktitle=iros,
  pages={4194--4198},
	year={2021},
}

@inproceedings{jiang2022iros-pumalio,
  title={{A LiDAR-inertial odometry with principled uncertainty modeling}},
  author={Jiang, Binqian and Shen, Shaojie},
  booktitle=iros,
  pages={13292--13299},
  year={2022},
}

@article{wang2023ral-swlio,
  title={{SW-LIO: A sliding window based tightly coupled LiDAR-inertial odometry}},
  author={Wang, Zelin and Liu, Xu and Yang, Limin and Gao, Feng},
  journal=ral,
  volume={8},
  number={10},
  pages={6675--6682},
  year={2023},
}

@article{xu2022tro-fastlio2,
  title={{FAST-LIO2: Fast direct LiDAR-inertial odometry}},
  author={Xu, Wei and Cai, Yixi and He, Dongjiao and Lin, Jiarong and Zhang, Fu},
  journal=tro,
  volume={38},
  number={4},
  year={2022},
  pages={2053--2073},
}

@article{vizzo2023ral-kissicp,
  title={{KISS-ICP: In defense of point-to-point ICP--simple, accurate, and robust registration if done the right way}},
  author={Vizzo, Ignacio and Guadagnino, Tiziano and Mersch, Benedikt and Wiesmann, Louis and Behley, Jens and Stachniss, Cyrill},
  journal=ral,
  volume={8},
  number={2},
  pages={1029--1036},
  year={2023},
}

@inproceedings{shan2020iros-liosam,
  title={{LIO-SAM: Tightly-coupled LiDAR inertial odometry via smoothing and mapping}},
  author={Shan, Tixiao and Englot, Brendan and Meyers, Drew and Wang, Wei and Ratti, Carlo and Rus, Daniela},
  booktitle=iros,
  year={2020},
  pages={5135--5142},
}

@inproceedings{chen2023icra-dlio,
  title={{Direct LiDAR-inertial odometry: Lightweight LIO with continuous-time motion correction}},
  author={Chen, Kenny and Nemiroff, Ryan and Lopez, Brett T},
  booktitle=icra,
  pages={3983--3989},
  year={2023},
}

@article{chen2024ral-iglio,
  title={{iG-LIO: An incremental GICP-based tightly-coupled LiDAR-inertial odometry}},
  author={Chen, Zijie and Xu, Yong and Yuan, Shenghai and Xie, Lihua},
  journal=ral,
  volume={9},
  number={2},
  pages={1883--1890},
  year={2024},
}

@article{ebadi2023tro-darpa-slam,
  title={{Present and future of SLAM in extreme environments: The DARPA SubT challenge}},
  author={Ebadi, Kamak and Bernreiter, Lukas and Biggie, Harel and Catt, Gavin and Chang, Yun and Chatterjee, Arghya and Denniston, Christopher E and Desch{\^e}nes, Simon-Pierre and Harlow, Kyle and Khattak, Shehryar and others},
  journal=tro,
  volume={40},
  pages={936--959},
  year={2023},
}

@article{tuna2025tfr,
  title={{Informed, constrained, aligned: A field analysis on degeneracy-aware point cloud registration in the wild}},
  author={Tuna, Turcan and Nubert, Julian and Pfreundschuh, Patrick and Cadena, Cesar and Khattak, Shehryar and Hutter, Marco},
  journal=tfr,
  volume={2},
  pages={485--515},
  year={2025},
}

@inproceedings{zhang2016icra-zhang,
  title={{On degeneracy of optimization-based state estimation problems}},
  author={Zhang, Ji and Kaess, Michael and Singh, Sanjiv},
  booktitle=icra,
  pages={809--816},
  year={2016},
}

@inproceedings{hinduja2019iros-hinduja,
  title={{Degeneracy-aware factors with applications to underwater SLAM}},
  author={Hinduja, Akshay and Ho, Bing-Jui and Kaess, Michael},
  booktitle=iros,
  pages={1293--1299},
  year={2019},
}

@article{tuna2023tro-xicp,
  title={{X-ICP: Localizability-aware LiDAR registration for robust localization in extreme environments}},
  author={Tuna, Turcan and Nubert, Julian and Nava, Yoshua and Khattak, Shehryar and Hutter, Marco},
  journal=tro,
  volume={40},
  pages={452--471},
  year={2023},
}

@article{chung2024ral-nvliom,
  title={{NV-LIOM: LiDAR-inertial odometry and mapping using normal vectors towards robust SLAM in multifloor environments}},
  author={Chung, Dongha and Kim, Jinwhan},
  journal=ral,
  year={2024},
  volume={9},
  number={11},
  pages={9375--9382},
}

@article{yang2024sensors-naloam,
  title={{NA-LOAM: Normal-based adaptive LiDAR odometry and mapping}},
  author={Yang, Fengli and Li, Wangfang and Zhao, Long},
  journal={IEEE Sensors Journal},
  year={2024},
  volume={24},
  number={19},
  pages={30715--30725},
}

@inproceedings{huang2024iros-lalio,
  title={{LA-LIO: Robust localizability-aware LiDAR-inertial odometry for challenging scenes}},
  author={Huang, Junjie and Zhang, Yunzhou and Xu, Qingdong and Wu, Song and Liu, Jun and Wang, Guiyuan and Liu, Wei},
  booktitle=iros,
  pages={10145--10152},
  year={2024},
}

@article{chen2022ral-dlo,
  title={{Direct LiDAR odometry: Fast localization with dense point clouds}},
  author={Chen, Kenny and Lopez, Brett T and Agha-mohammadi, Aliakbar and Mehta, Ankur},
  journal=ral,
  volume={7},
  number={2},
  pages={2000--2007},
  year={2022},
}

@inproceedings{lim2023ur-adalio,
  title={{AdaLIO: Robust adaptive LiDAR-inertial odometry in degenerate indoor environments}},
  author={Lim, Hyungtae and Kim, Daebeom and Kim, Beomsoo and Myung, Hyun},
  booktitle=ur,
  pages={48--53},
  year={2023},
}

@article{lee2024ral-genzicp,
  title={{GenZ-ICP: generalizable and degeneracy-robust LiDAR odometry using an adaptive weighting}},
  author={Lee, Daehan and Lim, Hyungtae and Han, Soohee},
  journal=ral,
  year={2024},
  volume={10},
  number={1},
  pages={152--159},
}

@inproceedings{gelfand2003threedim-degeneracy,
  title={{Geometrically stable sampling for the ICP algorithm}},
  author={Gelfand, Natasha and Ikemoto, Leslie and Rusinkiewicz, Szymon and Levoy, Marc},
  booktitle=threedim,
  pages={260--267},
  year={2003},
}

@book{joseph1968filter-joseph,
  title={{Filtering for Stochastic Processes with Applications to Guidance}},
  author={Bucy, Richard S and Joseph, Peter D},
  year={1968},
  publisher={NY, USA: Interscience Publishers}
}

@bookchapter{schmidt1966acs-schmidt,
  title={{Application of state-space methods to navigation problems}},
  author={Schmidt, Stanley F},
  booktitle={Advances in Control Systems},
  pages={293--340},
  year={1966},
  publisher={Elsevier}
}

@inproceedings{besl1992point-to-point,
  title={{Method for registration of 3-D shapes}},
  author={Besl, Paul J and McKay, Neil D},
  booktitle={Sensor Fusion IV: Control Paradigms and Data Structures},
  volume={1611},
  pages={586--606},
  year={1992},
  organization={{SPIE}}
}

@inproceedings{rusinkiewicz2001point-to-plane,
  title={{Efficient variants of the ICP algorithm}},
  author={Rusinkiewicz, Szymon and Levoy, Marc},
  booktitle=threedim,
  pages={145--152},
  year={2001},
}

@book{cheney1998condition-number,
  title={{Numerical Mathematics and Computing}},
  author={Cheney, Ward and Kincaid, David},
  year={1998},
  publisher={International Thomson Publishing}
}
\bibliographystyle{URL-IEEEtrans}

\end{document}